# Neural Network Predictive Controller for Grid-Connected Virtual Synchronous Generator

Sepehr Saadatmand, Mohammad Saleh Sanjarinia, Pourya Shamsi, Mehdi Ferdowsi, and Donald C. Wunsch
Department of Electrical and Computer Engineering
Missouri University of Science and Technology
sszgz@mst.edu, mswvq@mst.edu, shamsip@mst.edu, ferdowsi@mst.edu, dwunsch@mst.edu

*Abstract*—In this paper, a neural network predictive controller is proposed to regulate the active and the reactive power delivered to the grid generated by a three-phase virtual inertia-based inverter. The concept of the conventional virtual synchronous generator (VSG) is discussed, and it is shown that when the inverter is connected to non-inductive grids, the conventional PI-based VSGs are unable to perform acceptable tracking. The concept of the neural network predictive controller is also discussed to replace the traditional VSGs. This replacement enables inverters to perform in both inductive and non-inductive grids. The simulation results confirm that a well-trained neural network predictive controller illustrates can adapt to any grid impedance angle, compared to the traditional PI-based virtual inertia controllers.

*Index Terms*-- Grid connected inverters, neural network predictive controller, optimal control, virtual synchronous generator

## I. INTRODUCTION

Environmental pollution and energy crises are encouraging the penetration of distributed generators (DG). The main portion of DGs are renewable energy sources (RESs) such as wind turbines and photovoltaics. Most of these RESs are connected to the grid through a three-phase inverter. The grid current control is the traditional method for controlling the power penetration of the inverter-based DGs. To synchronize the inverter with the grid, this method typically uses a phase-locked loop (PLL). Moreover, a controller is used in order to regulate the reactive power and the active power delivered to the grid. Typically, this controller is a conventional PI. The most relevant disadvantageous of this method have been identified. First, the system inertia decreases by fast response and inertia less controllers. Furthermore, the current control inverters are unable to perform as a grid and function in standalone mode [1], [2].

To overcome the traditional controller drawbacks, numerous solutions have been proposed. To present the 'sync' and 'inertia' mechanism of SGs to inverters, a novel control named the virtual synchronous generator (VSG) has been proposed by some scholars. Hence, this method imitates the behavior of the synchronous generators, the power oscillations and the stability of the power system are improved. The traditional method to implement VSGs is to apply a conventional PI to control the inverter voltage. By controlling the inverter voltage the reactive power tracks its reference. To track the active power reference the virtual inertia equation is used to set the phase angle. This controls both the frequency and the voltage in a decoupled method. Applying decoupled controller is a good approximation for inductive grid. However, if the inverter is connected to the grid via a non-inductive line, this method does not function properly [3]-[5]. The predictive controller can be applied to an optimal control problem if the model of the system is known. Nonetheless, when it comes to uncertainties this controller is not powerful enough. As a case in point, the line that connects the inverter to the grid might change, then to make sure that the controller functions accurately, the system model needs to be revised [6], [7]. To resolve the issue, a predictive control based on neural networks has been studied [8]. This method has been studied in different applications, such as adaptive automatic generation control [9], grid-connected DG inverters [10], direct-drive wind turbine generators [11], and transient-following control of active power filters [12].

The main contribution of this paper is to report our study in the implementation of a neural network predictive controller (NNPC) for virtual inertia-based grid-connected three-phase inverters. First, a brief review of the VSG concept and the virtual inertia controller is presented in Section II. Section III introduces the neural network predictive control for VSGs, and it explains how to implement and train the neural networks. The performance of the proposed NNPC controller is evaluated in Section IV. Finally, the conclusions are presented in Section V.

## II. PRINCIPLE AND MODEL OF THREE-PHASE INVERTERS

Figure 1 illustrates the circuit diagram of a three-phase grid-connected inverter. In this figure, $V_{dc}$ is the DC voltage storage connected to the inverter. Hence, the inverter can route power from the dc source to inject additional power to the grid during transients (not applicable to solar photovoltaic inverters. The switching nature of inverters causes the high frequency harmonics. To connect these inverters to the grid, a low-pass filter is needed to eliminate the switching frequency. A second-order LC filter is applied to perform as the low-pass filter. In this filter, $X_{F\_L}$ and $X_{F\_C}$ are the inductor reactance and the capacitor reactance, respectively. This low-pass filter allows the grid frequency (50 Hz /60 Hz) to pass, and it filters the high frequency switching harmonics. A grid-connected inverter can also provide the demanded power for a local load shown with $R_{ld}$ and can be connected to the grid via a three-phase circuit breaker. In this figure $X_L$ and $R_L$ are the reactance and the resistance of the line. The grid voltage is assumed as the reference voltage with the value of $V_g$ and the phase angle of zero. The filtered output current and the output voltage of the inverter can be measured and fed to the control unit to calculate the inverter voltage ($E$) and the phase angle of the inverter voltage ($\theta$). Finally, a PWM unit, converts these values to three pulse signals to drive the inverter switches.

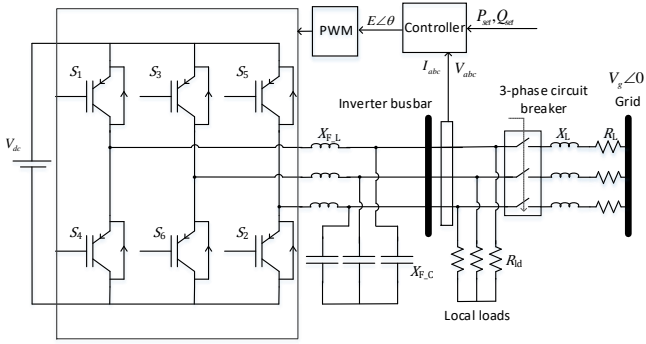

Figure 1. A conventional grid-connected three-phase inverter

Various studies have defined different types of controller for an inverter. Current-controlled method (such as, current $H\infty$ repetitive control, current proportional-resonant control, current proportional–integral control, and current deadbeat predictive control) and voltage-controlled method (such as voltage $H\infty$ repetitive control and Synchronverters) have been studied. In addition, the advantages and the drawbacks of these controllers have been determined [13]. In this paper, the goal is to improve the traditional synchronverters to offset its weak points.

### A. VSG controller

The goal for the conventional current-controlled inverters is to inject the maximum power from the associated RES to the grid. This method is useful while the portion of these inverters is negligible compared to the grid size. However, a voltage-controlled inverter responding similarly to the traditional generators is preferred when an inverter is connected to a weak grid or a microgrid, or operates in standalone mode. In this section, a control method is presented to mimic the response of the synchronous generator. By applying this method, the reactive power and the real power delivering to the grid can be automatically shared by using the traditional method of the frequency-droop and the voltage-droop control.

The mechanical equation for the machine can be written as:

$$J\ddot{\theta} = T_m - T_e - D_p \Delta \dot{\theta} \quad (1)$$

where $J$, $T_e$, $T_m$, $D_p$, and $\theta$ are the moment of inertia of the rotating parts, the electromagnetic toque, the mechanical torque, a damping factor, and the phase angle of the rotor, respectively. Assuming that the inverter operates around the reference angular velocity of $w_o$, the mechanical and the

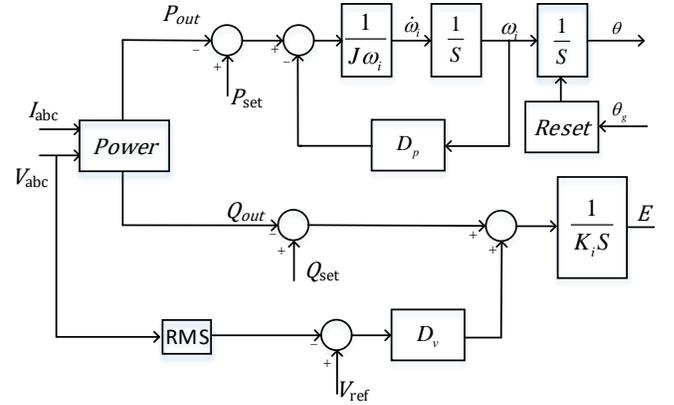

Figure 2. VSG controller block diagram

electrical torque can be replaced. In other words, Equation (1) can be rewritten as follows:

$$P_{set} - P_{out} = J\omega_i \dot{\omega}_i + D_p \Delta \omega_i \quad (2)$$

$$w_i = \dot{\theta}$$

$$P_{set} = T_m/\omega_i$$

$$\Delta \omega_i = \omega_i - \omega_{ref}$$

$$P_{out} = T_e/\omega_i.$$

where $P_{in}$, $P_{out}$, $\omega_i$, and $\omega_{ref}$ are the input power to synchronous generator, the electric output power, the angular velocity of the rotor, and the reference angular velocity, respectively.

The command signal to the inverter includes two parts. First, it needs the inverter voltage magnitude ($E$). Secondly, it needs the inverter voltage phase with respect to the grid ($\delta$). In order to compute $E$, the electrical output power can be computed by measuring the inverter voltage signals and the current signals injected into the grid. Having all the parameters in (2), $\omega_i$ can be computed at each control cycle. Then, the mechanical phase can be calculated by integrating this frequency as follows:

$$\theta = \int \omega_i \cdot dt$$

In order to control the inverter's output voltage, a reactive power controller with a voltage-droop is utilized. Applying a voltage-droop and an integrator-controller generates the RMS/peak value of the voltage as follows:

$$E = \frac{1}{K_i} \int \Delta Q \cdot dt - D_v \Delta V$$

where $K_i$ and $D_v$ are the integrator coefficient and the voltage droop, respectively. The inverter reactive power tracking error is given by $\Delta Q = Q_{set} - Q_{out}$, and the inverter voltage tracking error is given by $\Delta V = V_{ref} - V_i$. The reference reactive power for the inverter is set to $Q_{set}$ and the inverter output reactive power can be computed by a power meter block. The variable $V_i$ is the inverter output voltage, and $V_{ref}$ is the reference voltage for the inverter. Figure 2 shows the block diagram of the proposed virtual synchronous generator.

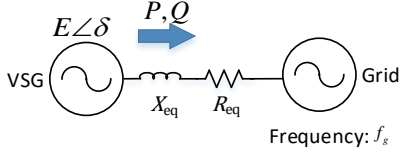

Figure 3. Averaged circuit model of a grid-connected inverter

### B. Power flow equation for grid-connected inverters

The proposed VSG averaged model can be derived based on a voltage source as shown in Figure 3. In this model, the local load is ignored. In the figure, $X_{eq}$ is the equivalent reactance per phase, $R_{eq}$ presents the equivalent resistance, and $Z_{eq}$ is the equivalent impedance per phase (line and filter) given as $Z_{eq} = jX_{eq} + R_{eq}$. The reactive and real power generated by the inverter and delivered to the grid can be computed as

$$Q = \frac{1}{2}\left[\left(\frac{E^2}{Z_{eq}^2} - \frac{EV\cos\delta}{Z_{eq}^2}\right)X_{eq} - \frac{EV}{Z_{eq}^2}R_{eq}\sin\delta\right]$$

$$P = \frac{1}{2}\left[\left(\frac{E^2}{Z_{eq}^2} - \frac{EV\cos\delta}{Z_{eq}^2}\right)R_{eq} + \frac{EV}{Z_{eq}^2}X_{eq}\sin\delta\right]$$

where $Q$ and $P$ are the delivered reactive and real power (per phase), $V$ is the grid voltage peak value, $E$ is the equivalent inverter voltage peak value, and $\delta$ is the phase angle between the grid voltage and the inverter voltage. For an inductive equivalent impedance (i.e. $X_{eq} \gg R_{eq}$) the active and reactive power can be estimated as

$$Q \approx \frac{E}{2X_{eq}}(E - V\cos\delta) \qquad (3)$$

$$P \approx \frac{EV}{2X_{eq}}\sin\delta. \qquad (4)$$

Generally, the inverter power angle $\delta$ is small, and $\sin\delta$ can be approximated by $\delta$, and $\cos\delta$ can be approximated by 1. Therefore, (3) and (4) can be written as

$$Q \approx \frac{E}{2X_{eq}}(E - V) \qquad (5)$$

$$P \approx \frac{EV}{2X_{eq}}\delta \qquad (6)$$

Equation (5) and (6) clarify that in inductive grids, the reactive power is proportional to the inverter voltage and the real power is proportional to the inverter power angle. In this case, the conventional VSG controller performance is acceptable; nonetheless, in low voltage grids that are mostly resistive or semi-resistive this assumption is no longer valid. In other words, Q is proportional to both the phase angle and the voltage magnitude. In order to turn the reactive power controller for non-inductive grids, all the parameters of the system model need to be known to make it possible to design an acceptable reactive power controller. However, in the power system, the inverter might face uncertainties such as line impedance changes, or nonlinear behaviors (e.g. transformer saturation) in the electrical element, that alter the reactive power equation. In this paper, an adaptive dynamic controller capable of adjusting its parameters is used to find the optimal solution and the results are compared with the conventional controller performance.

### III. NEURAL NETWORK MODEL PREDICTIVE CONTROLLER

In this section, the principle of the neural network predictive control is presented. First, a brief overview of the concept of the model predicative control is presented, and then the limitation of this method is explained. Following, the concept of the neural network predictive controller is explained.

### A. Model predictive controller (MPC)

Assume that the mathematic equation of the dynamic model of a system is known. The state space model of this system can be written as

$$\dot{X}_c(t) = F_c(X_c(t), U_c(t)) \qquad (7)$$

$$Y_c(t) = G_c(X_c(t), U_c(t)) \qquad (8)$$

where $X_c(t)$, $U_c(t)$, and $Y_c(t)$ are the state vector, the input vector, and the output vector, respectively. The function $F_c(\cdot)$ determines the derivative of state vector in terms of state vector and the control input vector. Moreover, the function $G_c(\cdot)$ defines the output vector in terms of the state vector and input vector. In (7) and (8), the c index states the continuous domain. These equations can be rewritten in discontinuous domain as follows,

$$X(k+1) = F(X(k), U(k))$$

$$Y(k) = G(X(k), U(k)), \qquad k = 1, 2, 3, \ldots$$

Based on the discontinuous state space model, it is possible to predict the output if the current state is known and the future control is determined. The main goal of the predictive control is to adjust the future control vector so that the output vector follows the desired trajectory as close as possible. In typical form of the model predictive controller, the control vector needs to be defined to minimize the following cost function in a specific horizon of time:

$$J(k) = \sum_{i=1}^{N_H}\|R(k+i) - Y(k+i)\|^2 + \gamma\sum_{i=1}^{N_H}\|\Delta U(k+i-1)\|^2$$

where $N_H$ is the prediction horizon, $R(\cdot)$ is the reference signal vector, and $\gamma$ is the weight factor for the control vector signal. Every control signal might face constraints. This equation can be analytically solved, but the exact model of the system needs to be known.

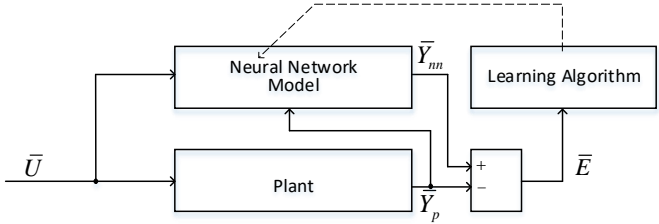

Figure 4. Neural network training block diagram

## B. Neural network predictive control (NNPC)

As mentioned, in order to implement the model predictive controller, the system model needs to be known. Nonetheless, there are systems in which the model is not known or the system model parameters are uncertain. In these cases, a neural network predictive control can conquer this issue by modeling the system via a neural network. In other words, instead of using the state space model to predict the future outputs, a neural network is used to complete this task. Figure 4 illustrates the training procedure of the neural network. In this figure, $\bar{U}$, $\bar{Y}_p$, $\bar{Y}_{nn}$, and $\bar{E}$ are the control vector, the plant output vector, the neural network output vector, and the error vector output, respectively. The neural network is fed with the current output vector and the control vector, and it predicts the next state output vector. Comparing the output vector of the neural network and the plant output vector, the error vector can be generated. By feeding the error vector to a learning algorithm, such as back propagation, the neural network is trained.

Figure 5 illustrates the model neural network, which is a fully connected multi-layer forward network. This network includes multiple hidden layers, and each hidden layer includes multiple nodes. The input to the neural network is the control input and the current plant output. The neural network output is the prediction for the next step output vector. The control input vector in this paper only includes one element, which is the voltage magnitude of the inverter. The neural network/plant output vector includes the inverter reactive and real power, the reactive power error, the real power error, the frequency error, and the inverter phase angle. The proposed neural network in this paper includes two hidden layers and seven nodes per each hidden layer. A set of data should be collected while the plant is operating by the conventional PI-based virtual inertia controller. Then, this set of data is used to train the neural network in the batch mode.

After training the neural network, the general model predictive controller method can be used to optimize the cost function for the defined horizon. Figure 6 illustrates the block diagram of the neural network predictive controller. As shown in this figure, the controller includes two blocks: the model neural network and the optimization block. In this figure, $\bar{N}_{input}$ is the neural network input vector. This vector includes the previous neural network outputs and the control vector. To minimize $J(\cdot)$, the optimization block generates a set of control vectors and constructs the neural network input vector, and then the optimal control vector ($\bar{U}$) is fed to the plant. In this paper, the time horizon is one second and the time step is one millisecond.

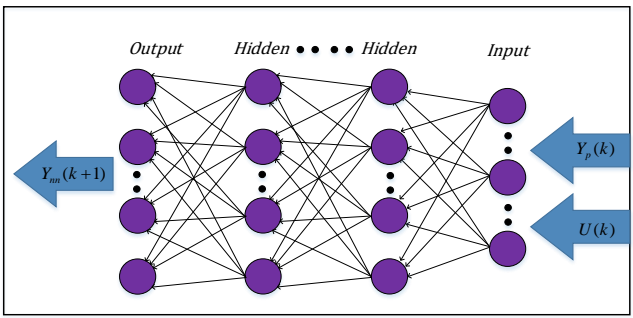

Figure 5. A fully connected neural network model with multiple hidden layers

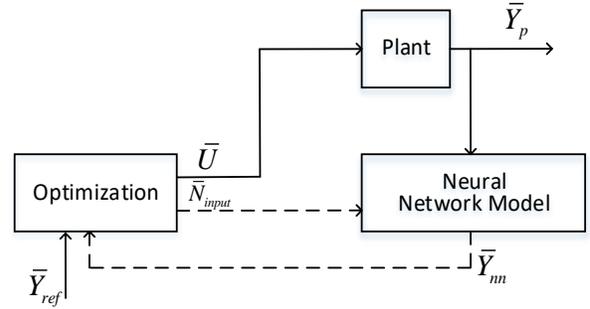

Figure 6. The block diagram of neural network predictive controller

## IV. PERFORMANCE EVALUATION OF THE TRAINED NNPC VIRTUAL INERTIA−BASED CONTROLLER

Figure 7 illustrates the block diagram of the proposed NNPC VSG controller. In the proposed controller the integral voltage droop controller is replaced with the neural network predictive controller.

TABLE I
SYSTEM PARAMETERS USED IN SIMULATION

| Parameter | Value | Unit |
|---|---|---|
| DC voltage | 250 | V |
| AC line voltage | 110 | V |
| AC frequency | 60 | Hz |
| Moment of inertia | 0.1 | Kg.m2 |
| Frequency droop | %4 | -- |
| Inverter power rating | 5 | kW |
| Inductive line | | |
| Filter inductance | 1e-6 | H |
| Line inductance | 1e-4 | H |
| Line resistance | 1e-2 | Ω |
| Resistive line | | |
| Filter inductance | 1e-6 | H |
| Line inductance | 1e-6 | H |
| Line resistance | 5e-1 | Ω |
| NNPC parameters | | |
| Time horizon | 1 | sec |
| Sampling time | 1e-3 | sec |
| Hidden layer | 2 | -- |
| Node per hidden layer | 7 | -- |
| Control weight factor | 0 | -- |

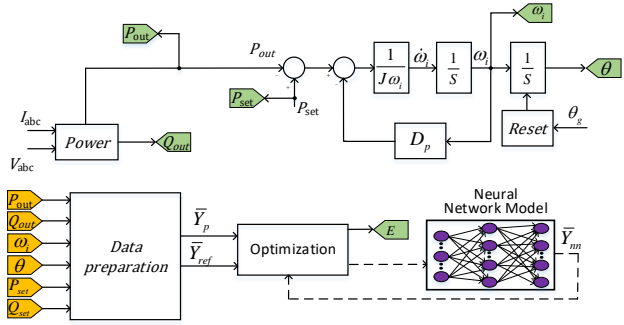

Figure 7. Neural network predictive controller applied to the virtual synchronous generator.

In order to train the neural network, the VSG is controlled by the traditional PI voltage-droop controller for 1000 s. At the time that the system reaches to the steady state vicinity, the reactive power and the real power change randomly to generate new data. All the data, including the active and the reactive power output, the reference values, the magnitude of the inverter voltage, the phase angle of the inverter and the virtual frequency, is stored. Using the backpropagation, the neural network is trained. To verify the proposed controller, NNMP controller is implemented for both inductive and resistive grids. The system parameters are listed in Table I. In order to solve the optimization problem, that the voltage changes are assumed to be chosen from a set defined as

$IN_{set} = \{-5, -1, -0.2, -0.04, 0, 0.04, 0.2, 1, 5\}$,

and the optimization block computes $J$ for each input, finds the optimal control signal, and feeds it to the plant.

### A. Inductive grid

As mentioned in Section II, in inductive grid connection, the typical VSG assumption is valid. It means that, the reactive power is proportional to the voltage magnitude, and the active power is proportional to the inverter phase angle. In this part, the proposed neural network predictive controller is implemented to regulate a VSG-based three-phase inverter connected to the grid with an inductive line. The typical VSG controller uses the swing equation to compute the inverter phase angle, and uses the reactive power error to compute the voltage magnitude. Similarly, the neural network predictive controller uses the swing equation to compute the inverter phase angle; however, it uses both the active power error and the reactive power error to generate the inverter voltage magnitude. Figure 7 illustrates the comparison between the traditional VSG controller and the proposed neural network predictive controller. As was expected, the performance of the PI-based VSG is acceptable; nevertheless, the performance of the neural network controller includes less accumulative error during the time horizon.

### B. Resistive grid

Figure 8 depicts the comparison between the neural network predictive controller and the conventional PI-based VSG inverter, connected to the grid with a resistive line. As discussed in Section II, the assumption that the reactive and active power are proportional to the inverter voltage magnitude

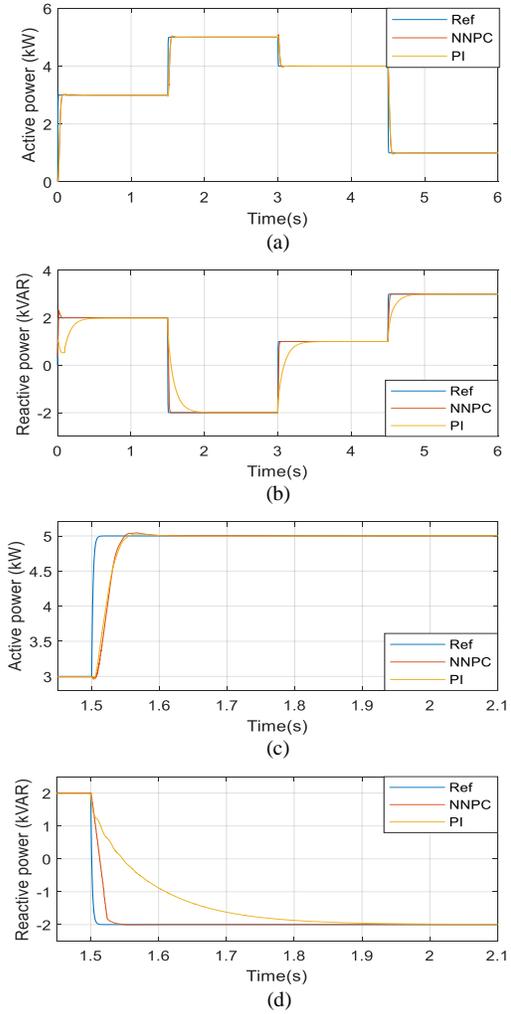

Figure 7. NNPC performance for VSG connected to the inductive grid (a) active power, (b) reactive power, (c) active power (zoomed in), (d) reactive power (zoomed in)

and phase angle, respectively, is no longer valid in resistive grids. Consequently, the conventional controller, which uses the reactive power error to regulate the inverter magnitude, does not function properly. The inverter voltage magnitude not only changes the reactive power, but also alters the active power as well. In this part, a tuned PI controller is also applied to control the inverter voltage magnitude. The input to this inverter is the combination of the active and the reactive power. However, the phase angle is still defined by the swing equation, which does not include the reactive power error directly. As was expected, the tuned PI controller performance is better than that of the typical PI; however, the overshoot and the settling time are not acceptable yet. Finally, the neural network predictive controller is applied to the virtual synchronous generator. The neural network-based nature of the NNPD controller enables itself to adjust the networks' weights through an offline learning at the beginning of the control process design to guarantee the performance necessities. As shown, the NNPC response has the smallest overshoot and the fastest settling time compared to PI-VSG and the tuned PI-VSG.

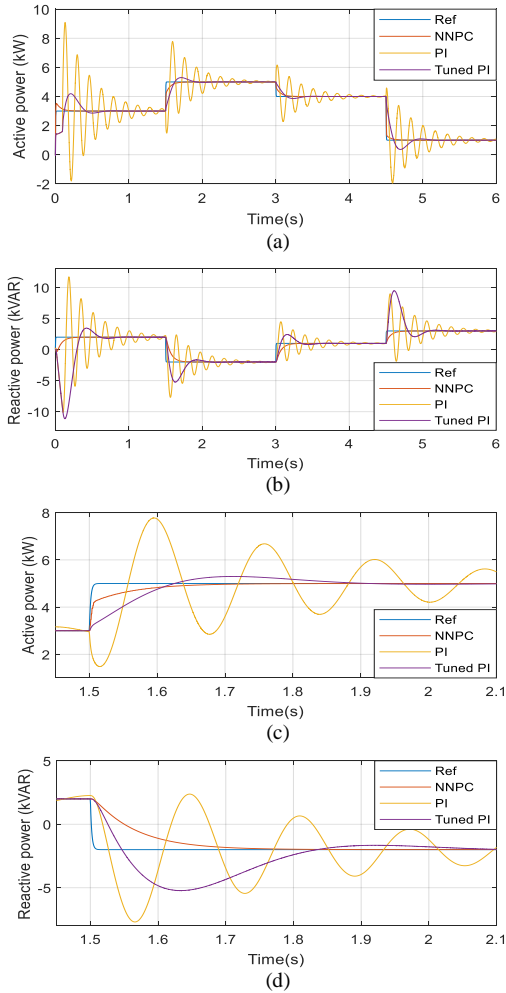

Figure 8. NNPC controller performance for VSG connected to the resistive grid. (a) active power, (b) reactive power, (c) active power (zoomed in), (d) reactive power (zoomed in)

## V. CONCLUSION

A droop-based current control with conventional PI controller typically controls these inverters, which has several drawbacks such as lack of the ability to work in standalone mode and stability issues. The VSG concept has been proposed to overcome these disadvantages. The VSGs are designed to be implemented in inductive grids. However, to implement conventional VSGs in non-inductive grids, the voltage controller block has to be redesigned. This paper presented the neural network predicative controller to regulate the inverter voltage in VSGs. The proposed controller is able to perform in inductive and non-inductive grid connection. It was shown by simulation that a well-trained neural network predictive controller considerably responds better than the PI-based and tuned PI-based VSGs. Particularly, by applying the NNPC in resistive connected inverters, the overshoot and the settling time are extremely reduced. Comparison results demonstrated the effectiveness of the proposed controller in both resistive and inductive grids.

## VI. ACKNOWLEDGMENT

This material is based upon work supported by the U.S. Department of Energy, "Enabling Extreme Fast Charging with Energy Storage", DE-EE0008449.

## VII. REFERENCES

[1] F. Blaabjerg, R. Teodorescu, M. Liserre, and A. Timbus, "Overview of Control and Grid Synchronization for Distributed Power Generation Systems," IEEE Transactions on Industrial Electronics, vol. 53, no. 5, pp. 1398–1409, 2006.

[2] A. K. Srivastava, A. A. Kumar, and N. N. Schulz, "Impact of Distributed Generations With Energy Storage Devices on the Electric Grid," IEEE Systems Journal, vol. 6, no. 1, pp. 110–117, 2012.

[3] H.-P. Beck and R. Hesse, "Virtual synchronous machine," in Proc. 9th Int. Conf. Elect. Power Qual. Utilisation, 2007, pp. 1–6

[4] S. D'Arco and J. A. Suul, "Equivalence of virtual synchronous machines and frequency-droops for converter-based microgrids," IEEE Trans. Smart Grid, vol. 5, no. 1, pp. 394–395, Jan. 2014.

[5] Ma, Jing, Yang Qiu, Yinan Li, Weibo Zhang, Zhanxiang Song, and James S. Thorp. "Research on the impact of DFIG virtual inertia control on power system small-signal stability considering the phase-locked loop." IEEE Transactions on Power Systems 32, no. 3 (2017).

[6] Kirk, D. E., Optimal Control Theory: An Introduction, Prentice–Hall, Englewood Cliffs, NJ, 1970, Chaps. 1–3.

[7] J. Rodriguez, P. Cortes, R. Kennel, and M. Kazrnierkowski, "Model predictive control -- a simple and powerful method to control power converters," 2009 IEEE 6th International Power Electronics and Motion Control Conference, 2009.

[8] O. Karahan, C. Ozgen, U. Hahci, and K. Leblebicioglu, "Nonlinear model predictive controller using neural network," Proceedings of International Conference on Neural Networks (ICNN97).

[9] D. Chen and L. Wang, "Adaptive automatic generation control based on gain scheduling and neural networks," 2017 IEEE Power & Energy Society General Meeting, 2017.

[10] H. Su-Fen, Y. Ling-Zhi, L. Ju-Cheng, Y. Zhe-Zhi, and P. Han-Mei, "Design of three-phase photovoltaic grid connected inverter based on RBF neural network," 2009 International Conference on Sustainable Power Generation and Supply, 2009.

[11] Y. F. Ren and G. Q. Bao, "Control Strategy of Maximum Wind Energy Capture of Direct-Drive Wind Turbine Generator Based on Neural-Network," 2010 Asia-Pacific Power and Energy Engineering Conference, 2010.

[12] J. Marks and T. Green, "Predictive transient-following control of shunt and series active power filters," IEEE Transactions on Power Electronics, vol. 17, no. 4, pp. 574–584, 2002.

[13] Q.-C. Zhong and T. Hornik, Control of Power Inverters in Renewable Energy and Smart Grid Integration. Hoboken, NJ, USA: Wiley/IEEE, 2013.